\documentclass[pdflatex,sn-mathphys-num]{sn-jnl}

\usepackage{graphicx}%
\usepackage{multirow}%
\usepackage{amsmath,amssymb,amsfonts}%
\usepackage{amsthm}%
\usepackage{mathrsfs}%
\usepackage[title]{appendix}%
\usepackage{xcolor}%
\usepackage{textcomp}%
\usepackage{manyfoot}%
\usepackage{booktabs}%
\usepackage{algorithm}%
\usepackage{algorithmicx}%
\usepackage{algpseudocode}%
\usepackage{listings}%
\usepackage{acro}
\usepackage{subcaption}
\usepackage{graphicx}
\usepackage{float}
\usepackage{lineno}
\usepackage{makecell}

\theoremstyle{thmstyleone}%
\theoremstyle{thmstyletwo}%

\theoremstyle{thmstylethree}%

\DeclareAcronym{mic}{
  short=MiC,
  long=Modular Integrated Construction,
}
\DeclareAcronym{bim}{
  short=BIM,
  long=Building Information Models,
}
\DeclareAcronym{vr}{
  short=VR,
  long=Virtual Reality,
}

\DeclareAcronym{top}{
  short=CRCUST\_Top,
  long=A fully automated safety monitoring system for \ac{mic} lifting on the construction top,
}

\DeclareAcronym{pc}{
  short=PC,
  long=Point Cloud,
}

\DeclareAcronym{gui}{
  short=GUI,
  long=Graphical User Interface,
}

\DeclareAcronym{sam}{
  short=SAM,
  long=Segment Anything Model,
}

\DeclareAcronym{dfl}{
  short=DFL,
  long=Distributed Focal Loss,
}

\DeclareAcronym{lidar}{
  short=LiDAR,
  long=Light Detection And Ranging,
}

\DeclareAcronym{sota}{
  short=SOTA,
  long=State Of The Arts,
}

\DeclareAcronym{iot}{
  short=IoT,
  long=Internet of Things,
}

\DeclareAcronym{uav}{
  short=UAV,
  long=Unmanned Aerial Vehicles,
}

\DeclareAcronym{yolo}{
  short=YOLO,
  long=You Only Look Once,
}

\DeclareAcronym{ai}{
  short=AI,
  long=Artificial Intelligence,
}

\DeclareAcronym{csf}{
  short=CSF,
  long=Cloth Simulation Filtering,
}

\DeclareAcronym{imu}{
  short=IMU,
  long=Inertial Measurement Unit,
}

\DeclareAcronym{ap}{
  short=AP,
  long=Average Precision,
}

\DeclareAcronym{map}{
  short=mAP,
  long=mean Average Precision,
}

\DeclareAcronym{iou}{
  short=IoU,
  long=Intersection over Union,
}

\begin{document}

\title[Bird's-eye view safety monitoring for the construction top under the tower crane]{Bird's-eye view safety monitoring for the construction top under the tower crane}


\author[1]{\fnm{Yanke} \sur{Wang}}\email{yankee.wann@gmail.com}

\author[1]{\fnm{Yu Hin} \sur{Ng}}\email{yhngad@connect.ust.hk}

\author[1]{\fnm{Haobo} \sur{Liang}}\email{hbliang@ust.hk}

\author[1]{\fnm{Ching-Wei} \sur{Chang}}\email{ccw@ust.hk}

\author*[1]{\fnm{Hao} \sur{Chen}}\email{danielchen16@hotmail.com}

\affil[1]{\orgdiv{Hong Kong Center for Construction Robotics}, \orgname{The Hong Kong University of Science and Technology}, \orgaddress{\street{Units 808 to 813 and 815, 8/F, Building 17W, Hong Kong Science Park}, \city{Pak Shek Kok}, \state{New Territories}, \country{Hong Kong SAR}}}


\abstract{The tower crane is involving more automated and intelligent operation procedure, and importantly, the application of automation technologies to the safety issues is imperative ahead of the utilization of any other advances. Among diverse risk management tasks on site, it is essential to protect the human workers on the workspace between the tower crane and constructed building top area (construction top) from the bird's-eye view, especially with Modular Integrated Construction (MiC) lifted. Also, the camera and Light Detection And Ranging (LiDAR) can capture abundant 3D information on site, which is however yet made the best use. Considering the safety protection for humans and tower cranes, we present an AI-based fully automated safety monitoring system for tower crane lifting from the bird’s-eye view, surveilling to shield the human workers on the construction top and avoid cranes' collision by alarming the crane operator. The system achieved a 3D data fusion for localization of humans and MiCs by integrating the captured information from camera and LiDAR. The state-of-the-art methods were explored and implemented into our proposed software pipeline coupled with the hardware and display systems. Furthermore, we conducted an analysis of the components in the pipeline to verify the accuracy and effectiveness of the involved methods. The display and visualization on the real site proved that our system can serve as a valuable safety monitoring toolkit on site.}




\maketitle

\section{Introduction}\
\label{sec:intro}
Tower cranes are essential on modern construction sites, facilitating the lifting and handling of heavy material. However, safety incidents related to tower cranes still occur frequently even though many governments propose restricted operation rules \cite{media2024new,scal2024lifting}. According to statistics \cite{ali2024tower}, China recorded nearly 125 tower cranes accidents in a single year, and USA witnessed 129 recorded construction crane accidents between 2011 and 2020. Singapore experienced six crane-related fatalities \cite{ku2024robotic} while there were three fatalities in Hong Kong SAR \cite{ali2024tower} in the year 2022. How to protect the human operators of tower cranes and workers on the construction site has become an intuitively urgent issue since recent accidents keep occurring and continuing annually as the Occupational Safety and Health Administration of USA reported \cite{accident2023usa}. Based on the knowledge above, increasing attentions are attracted by the safety monitoring of the tower cranes among a great mount of safety concerns \cite{ali2024tower,pazari2023enhancing,lam2023chapter,shen2021deep}. 

The previous research on the safety monitoring of tower cranes is more focused on monitoring the status of tower crane itself. For example, \cite{9281364} used the vision-based method to obtain the payload displacement. \ac{iot} technology was also used in \cite{guanghui2020research} to obtain real-time sensor operation data and dynamic structure of tower cranes. In the work of \cite{yang2020safety}, multisensory technology was applied for monitoring the real-time working status of tower cranes. In addition to monitoring the status of the tower crane itself, more researchers begin to take account the human-centered safety issue into site monitoring task. In such case, the deep learning methods are commonly used in the safety monitoring pipeline. For instance, \cite{pazari2023enhancing} implemented a \ac{yolo}v7 to monitor the working space of the workers. Besides, \cite{lam2023chapter} developed a vision-based pipeline to surveil the activities of the workers under the tower crane, e.g., wearing the helmet (or vest) or not. Moreover, the current safety monitoring of the tower crane is mainly based on images \cite{pazari2023enhancing,lam2023chapter}, which cannot provide depth information. However, the research about three-dimensional safety monitoring is relatively limited \cite{shen2021deep}. In terms of viewpoint for safety monitoring, the bird's eye view is more suitable for the safety monitoring of tower cranes, as the monitoring equipment is commonly installed on or over the tower crane that can cover almost the entire lifting process~\cite{higgins2023unlocking}.

Recently, \ac{mic} is afoot to predominate the construction site owing to its high efficiency, safety guarantee, productivity, and sustainable pre-fabrication \cite{polyu2021a}. \ac{mic} consists of producing building modules off-site, transporting them to the construction site, and assembling them with minimal on-site activities \cite{ali2024crane}. Although \ac{mic} has numerous benefits, it still carries various risks such as crane breakdown issues, inadequate data coordination, modular installation faults, modular production system failures, manual inspecting, and unwrapping among others \cite{khan2021systematic}. The safety issues are mostly focused on the process of fabrication, with concerns by 'fracture' or 'fall' as a result of the inherently unstable structure \cite{fard2017safety} in addition to external environmental and organizational factors \cite{song2022study}. Moreover, because of the large size of \ac{mic}, the visual blind area of the lifting is also greatly increased \cite{zhu2022crane}. Therefore, a safety monitoring system aiming to build up a safe and efficient \ac{mic} operation workspace for tower cranes is necessary.

After a detailed literature research, we raise the following questions,
\begin{itemize}
    \item How to define the safety concerns in the case of \ac{mic} lifting under the tower crane from a bird's-eye view?
    \item How to design an automated \ac{mic} safety lifting monitoring system based on the mentioned safety concerns?
    \item What \ac{ai} methods can be applied to the designed system and how do they perform?
    \item Whether the proposed system can be installed on the real site?
\end{itemize}
In response to the questions above, we propose an \ac{ai}-based 3D automated tower crane safety monitoring system from the bird's-eye view for \ac{mic} lifting. The monitoring area is the area between the constructed building top and the tower crane ("Construction Top" for simplicity). We aim to develop such system and apply it on a real construction site in Hong Kong. The challenges regarding the system development come from 1) the data acquisition is inadequate owing to the dangers presented on site; 2) the hidden dangers in the process of \ac{mic} lifting are complex and diverse, and how to define safety and insecurity is difficult; and 3) the accuracy and robustness of monitoring technologies are not guaranteed. To the best of our knowledge, there still not exists a sufficient implementation of fully 3D automated safety monitoring for crane lifting tasks of \ac{mic} from bird’s-eye view by using the state-of-the-art AI, especially in the case requiring the integration between camera and \ac{lidar}. We summarize our contributions as
\begin{itemize}
    \item In response to the safety concerns on the \ac{mic} site, definitions of what constitutes a safe site were provided;
    \item \ac{top}, involving hardware and information fusion between camera and \ac{lidar}, was proposed with AI capabilities;
    \item A \ac{top} dataset was collected on the real site for training and testing of the pipeline;
    \item The \ac{top} was applied to the real site, working continuously in a long-term manner for the verification of the pipeline.
\end{itemize}
The testing and verification results demonstrated that our system was adequate to perform the majority of safety monitoring tasks under our definitions. 

The remainder of this paper is structured as follows. Section \ref{sec:related} details the related work about safety monitoring on construction sites. Section \ref{sec:sys_dgn} introduces system design of the proposed safety monitoring system for the construction top under the tower crane. The implementation and results are described in Section \ref{sec:imple_res}, and conclusion is drawn in Section \ref{sec:conclusion}.

\begin{sidewaystable}
\caption{AI methods used in the safety monitoring on site.}
\label{tab:ai_safe}
\begin{tabular*}{\textheight}{@{\extracolsep\fill}lccccc}
\toprule
Method Type & References & 
Base Model &
Safety Issues (detected targets) & Data Source & Year \\
\midrule
 & \cite{neuhausen2018construction} & \makecell{Conventional image processing\\and machine learning} & Workers & Real-site & 2018 \\
\cmidrule{2-6}
& \cite{sutjaritvorakul2020data} &  RetinaNet & Workers & Synthetic & 2020\\
\cmidrule{2-6}
& \cite{neuhausen2020synthetic} & \makecell{Conventional image processing\\and machine learning} & Workers & Real-site and synthetic & 2020 \\
\cmidrule{2-6}
& \cite{yang2019safety} & Mask-RCNN & Workers, hazards & Real-site &2019\\
\cmidrule{2-6}
& \cite{chian2022dynamic} & CenterNet & Load zone & Real-site & 2022 \\
\cmidrule{2-6}
& \cite{bai2023automated} & YOLOv8-seg & \makecell{Construction machinery\\and operation surfaces} & Real-site & 2023 \\
\cmidrule{2-6}
Vision-based models & \cite{zhang2025state} & YOLOv8 & Tower cranes & Real-site & 2025 \\
\midrule
& \cite{xu2023virtual} & F-PointNet & MiC & Synthetic & 2023 \\
\cmidrule{2-6}
& \cite{wang2023lidar} & \makecell{Cloth Simulation\\Filtering (CSF)} & Tower cranes, lifting
objects & Real-site & 2023 \\
\cmidrule{2-6}
\ac{lidar}-based models & \cite{pei4743452active} & PointNet++ & Tower cranes, buildings & Lab-based & Preprint\\
\midrule
& \cite{price2021multisensor} & \makecell{Conventional image processing\\and machine learning} & \makecell{Load, obstacle,\\worker, tower cranes} & Real-site &  2021\\
\cmidrule{2-6}
& \cite{zhu2023physical} & Unity3D® game engine & Drone & Synthetic & 2023\\
\cmidrule{2-6}
\ac{lidar}-visiaon Fusion & \cite{shen2021deep} & Mask R-CNN & Objects, workers, crane truck & Real-site and synthetic & 2021 \\
\bottomrule
\end{tabular*}
\normalsize
\end{sidewaystable}

\section{Related work}
\label{sec:related}

\subsection{\ac{mic}-involved safety monitoring}
The lifting of \ac{mic} is most commonly operated by cranes on site, requiring a larger size and higher workload limit of cranes as well as more experienced site workers. This consequently exposes more challenges to the crane operation. In practice, research on \ac{mic}-involved safety monitoring on site needs to consider workforce safety \cite{mohandes2022occupational}, cranes' collision, and material damage. Diverse technologies, e.g., \ac{bim} \cite{chatzimichailidou2022using, darko2020building}, \ac{iot} \cite{zhai2019internet}, digital twin \cite{li2024real}, virtual reality \cite{zhang2021virtual} have been used in the safety management. The \ac{bim} is frequently applied in the risk assessment on site with modular construction \cite{chatzimichailidou2022using}, e.g., by designing a safer site layout scheme \cite{zhou2023location, yang2022bim} or crane operation planning with higher efficiency and safety \cite{bagheri2024automated}. 

In addition, the \ac{bim} technology can also be used with the integration of the other technologies. In \cite{li2024real}, researchers considered the payload of the tower crane and used digital twin technology to detect the workers under the fall zone of tower crane lifting \ac{mic} with the help of \ac{iot} and \ac{bim}. \cite{zhou2021customization} analyzed the potential risks in \ac{mic}, e.g., clash detection, by inputting the real-time 3D \ac{bim} to the virtual and digital environment. Seen from above, more and more pioneers are using more than one technology for the safety management and risk estimation. 

\subsection{Bird's-eye safety monitoring}
\label{sec:related_bird}
Although plenty of safety issues are concerning the construction stakeholders, monitoring the site safety from the bird's-eye view stands out as it provides a unique and compressive understanding of the regarding work place \cite{higgins2023unlocking}. This type of view can be established via two manners, i.e., a sensing system fixed on the body of the tower crane, and rotating with the crane jib or installed on the \ac{uav}. Two types of sensors are usually used, visual sensor and \ac{lidar}. Workers can be detected and tracked visually from bird's-eye view, e.g.,  \cite{neuhausen2018construction, neuhausen2020synthetic} did the detection from the view with 15 m above the workers. \cite{yang2019safety} installed a camera on the trolley of one boom tower crane, capturing the view surrounding the hook for measuring the safety distance between workers and site hazards. \cite{chian2022dynamic} used the existing camera installed on the top of tower crane and detected the load zone via vision-based method. \cite{bai2023automated} achieved the bird's-eye view by using a \ac{uav} with 20 m above the tower crane, aiming to monitor construction machinery and operation surfaces on site. Similarly, \cite{zhang2025state} used a \ac{uav} to monitor the movements of multiple cranes on site from the bird's-eye view. The details of the work mentioned above is outlined in Table \ref{tab:ai_safe}. The \ac{ai} methods used from the bird's-eye view are discussed in the next section according to the sensors installed in the risk management system.

\subsection{AI for the safe site}

As mentioned in the Section \ref{sec:related_bird}, the AI methods are applied according to the used sensor devices, i.e., vision-based and \ac{lidar}-based methods. This section aims to detail the literature research on visually servoing and \ac{lidar} enhanced safety monitoring under cranes as well as the \ac{sota} on the data fusion between vision and \ac{lidar} in the case of bird's-eye view under the tower crane. 

\subsubsection{Visual servoing safety monitoring}
The visual perception on site can be processed via the conventional image processing and machine learning methods or advanced deep learning-based models. In the scenarios with little environmental changes, the conventional image processing works stably and not computationally. For example, \cite{neuhausen2018construction} analyzed the histogram of the image and difference between two frames as well as Kalman filter for the worker detection and tracking. Moreover, they tested the pipeline both on real-site and synthetic data \cite{neuhausen2020synthetic}. These methods usually work in the limited workspace on site and rely on the user-initialized priors, e.g., the image threshold, object size information, target moving speed and range, etc. However, the aforementioned information shows dynamic changes on the real-time workspace, especially on the construction top area. Thus, the deep learning-based methods are potential to show their advantages in handling the changes of image brightness, color, and resolution as well as object scale and moving. The visual deep learning models commonly utilize convolutional neural networks for feature extraction, anchor-based mechanism for object detection, and fully convolutional neural networks for semantic segmentation. \cite{sutjaritvorakul2020data} followed the architecture of RetinaNet to do the detection of workers under crane from a load-view (similar to bird's-eye view). \cite{yang2019safety} used Mask-RCNN to predict the area and segments of workers and hazards. \cite{chian2022dynamic} adapted a CenterNet to detect the corners of load for defining the working area, and the categories of loads were pre-labeled. \cite{bai2023automated} improved \ac{yolo}v8-seg model for the segmentation of construction machinery and operation surfaces from remote sense images. Although the deep learning-based methods deliver excellent performance, they also suffer from the data issue since these methods require a large-scale dataset size to train a robust and accurate model. Apart from the machines and human workers on site, the safety of tower cranes has also been paid attention. \cite{zhang2025state} implemented a \ac{yolo}v8-based method for the skeleton detection of multiple tower cranes for avoiding the collision between cranes based on the safety distance assessment. The data collection strategy for the work above is also listed in Table \ref{tab:ai_safe}. 

\begin{figure}
\centering
\includegraphics[width=0.8\textwidth]{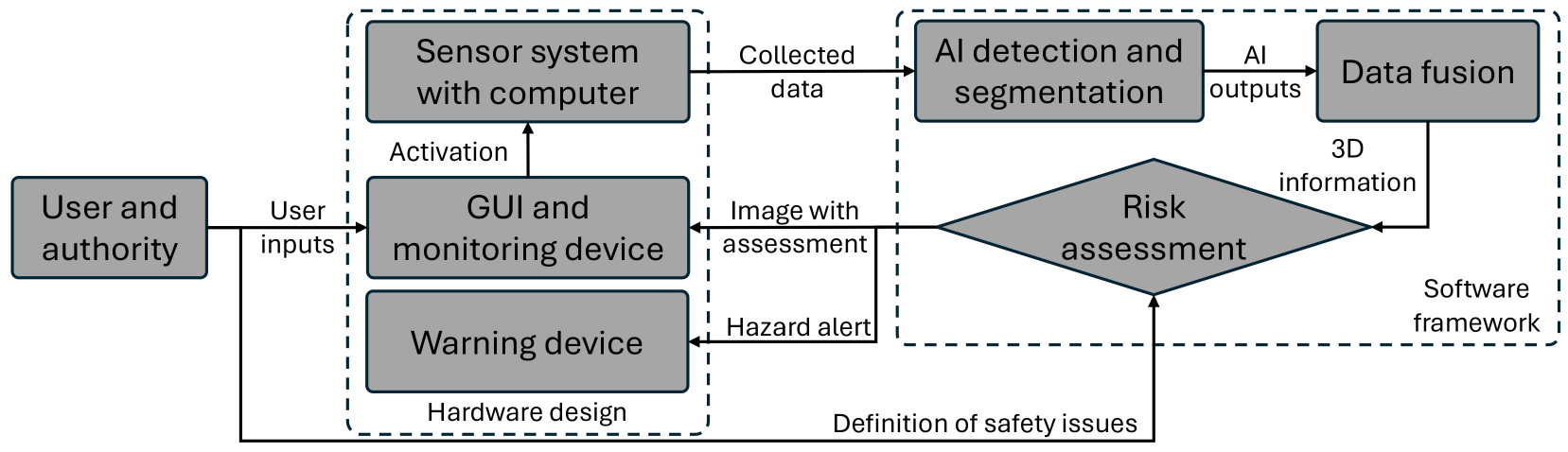}
\caption{The workflow of the deigned \ac{top}, including the components of the hardware design and software framework.}
\label{fig:system_flow}
\end{figure} 

\subsubsection{LiDAR enhanced safety monitoring}
The computer vision methods are sensitive to the image quality and might suffer from the accuracy for instant warning of risks \cite{li2024real}. Thus, an alternative device generating bird's-eye view is \ac{lidar}. However, only a very limited number of research is focused on the \ac{lidar}-based safety monitoring on site owing to its more challenging data collection procedure than 2D images. In response, \cite{xu2023virtual} offered a well-suited solution to the mentioned question in a virtual world. They simulated the lifting of different types of \ac{mic} on site for data generation and selected F-PointNet to do the 3D object detection. \cite{wang2023lidar} designed a \ac{lidar} system to do anti-collision between the lifting objects and tower crane via a \ac{csf}-based segmentation. 3D classification can also be used to generate anti-collision warning of tower cranes and buildings \cite{pei4743452active}, where a PointNet++ was applied to laboratory collected data. This section conveys a critical argument that research on bird's-eye view safety monitoring via \ac{lidar} is limited and still open for further investigation by scholars.

\subsubsection{\ac{lidar}-vision fusion}

This paper intends to implement the bird's-eye monitoring of lifting \ac{mic} based on the data fusion of 2D image and \ac{lidar}. Only a few research projects are related to this topic. In the context of related research, multiple sensors were integrated together to build a 3D visualization of the site with boom crane in \cite{price2021multisensor}. In that work, the camera was mounted at boom-tip, and a drone-based laser scanning was used to get the mapping of the environment. The above information were combined to generate a complete 3D model of the site. Besides, the boom angle sensor provided the pose estimation of the crane, so the work can fulfill obstacle and worker detection as well as monitor the crane working status. However, the update of the environment is not fully automated, and the drone-based point cloud generation is not real-time, so the application of the research to site with dynamical changes is still limited. Moreover, on the site with multiple drones, the collision between them is also an important safety issue, so \cite{zhu2023physical} simulated 4D drone-based virtual environment to evaluate the effects of flight parameters in risks under different scenarios. Despite not from bird's eye view, \cite{shen2021deep} tried to narrow the gap between \ac{ai} technologies and \ac{lidar}-vision fusion by generating pseudo point cloud data from 2D ground-view image. They modified mask R-CNN for 3D object detection from the generated point clouds and detected collisions between objects (workers) in the work zone of crane truck.

Based on the knowledge above, a research gap is clearly observed. In terms of safe tower crane site with \ac{mic} involved, the \ac{lidar} is still not widely used for safety monitoring, neither the \ac{lidar}-vision data fusion. Also, the \ac{sota} of \ac{ai} technologies still need more exploration and validation on the real site in the scenarios mentioned above. Thus, this study shows its significance by fulfilling the gap and providing the guidance in the regarding research.

\section{System design}
\label{sec:sys_dgn}
The objective of the designed system is to monitor the safety issues on the defined construction top workspace under the tower crane. The section provides the detailed solutions to the answers posed in Section \ref{sec:intro}, with the workflow as Fig.~\ref{fig:system_flow} visualizes. The user and authority provides user inputs (system settings, e.g., the thresholds and arguments) and definitions of safety issues according to the governmental regulations. Given the system settings, the \ac{gui} releases activation signal to the sensor system for data acquisition. The collected data are processed by the AI detection and segmentation methods with the outputs used for data fusion. The fused 3D information is then used for a risk assessment guided by the definition of safety issues. The assessed results are interpreted in two manners, i.e., an image with resulting information for a display on the \ac{gui} with a monitoring device and an alert of hazard via a warning device. The details are further discussed in this section, including the definitions of safety issues, coordinate system and pipeline, as well as the hardware and software design and the \ac{gui} used for a display. 

\begin{figure*}[h!]
\centering
\includegraphics[width=0.9\textwidth]{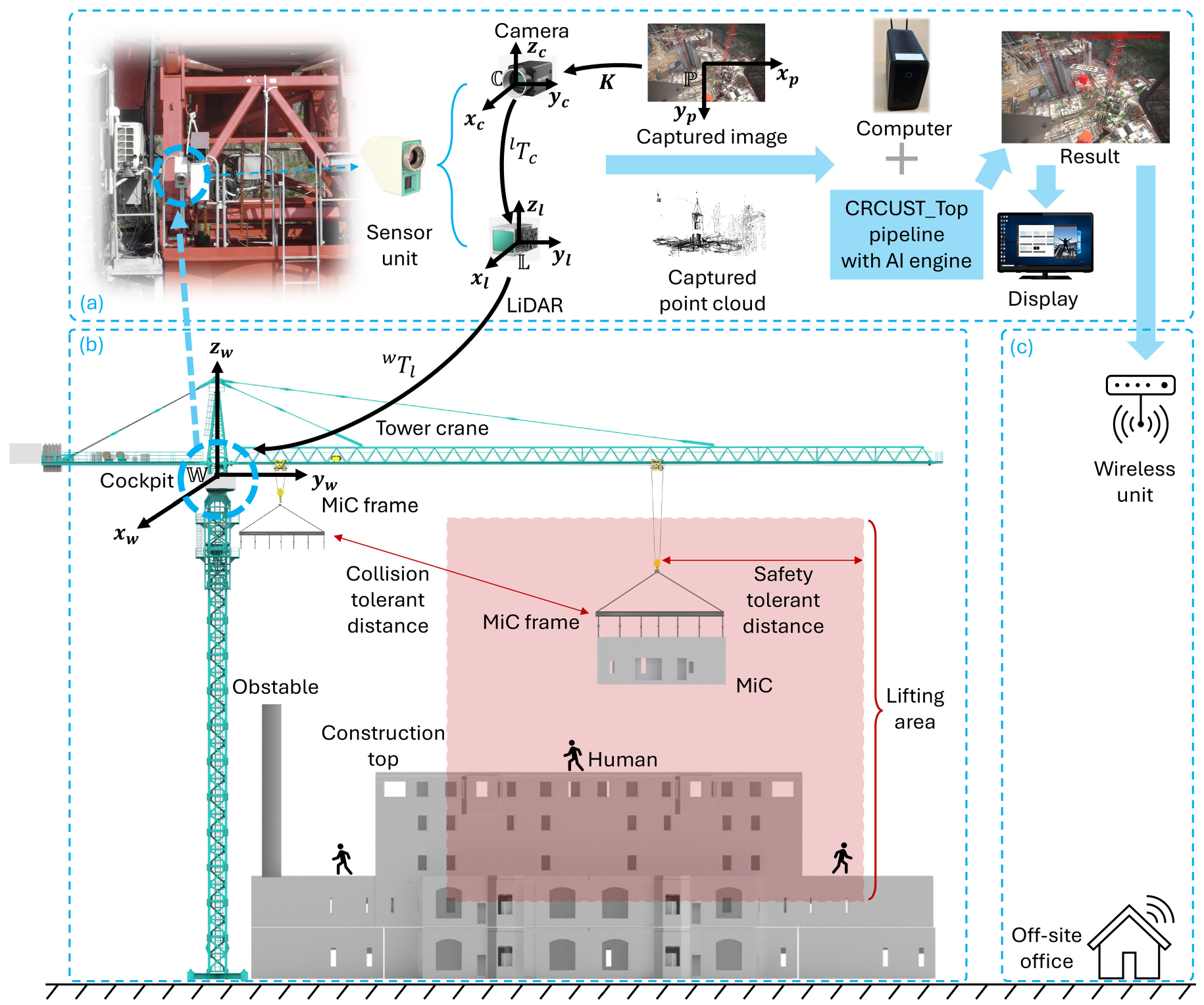}
\caption{The diagram showing the hardware design of \ac{top} and definition of the safety issues on the construction top. The hardware of the system installed on the cockpit contains a computer, a display, a sensor unit, and a wireless unit connecting with the off-site office. The safety issues include the protection of the humans presenting in the lifting area near to \ac{mic} on the construction top and avoidance of \ac{mic} frames' collision (cranes' collision).}
\label{fig:overall}
\end{figure*}


\begin{figure*}
\centering
\includegraphics[width=\textwidth]{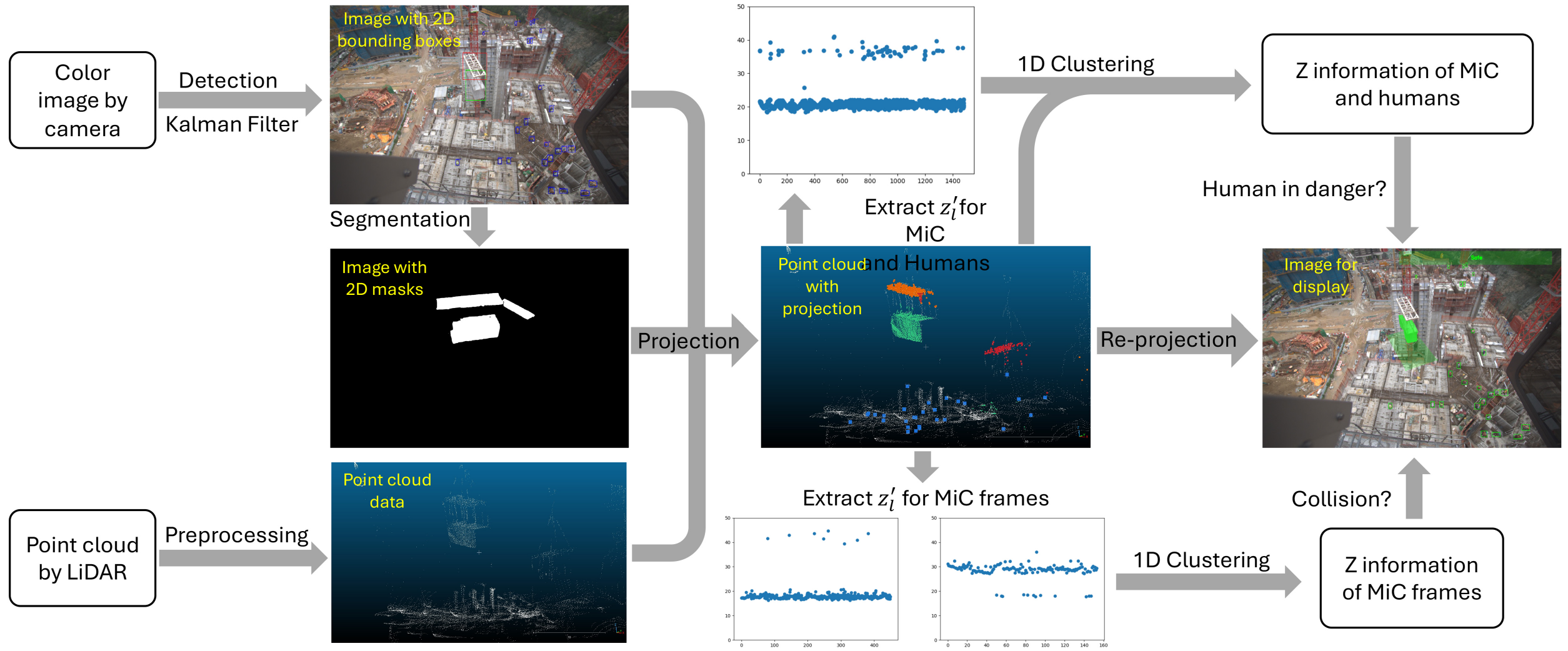}
\caption{The software pipeline of \ac{top}. The color image captured by the camera is processed by detection method and Kalman filter to obtain the 2D bounding boxes of humans, \ac{mic} and \ac{mic} frames, followed by a segmentation for 2D masks of them. The \ac{pc} data by \ac{lidar} are fused with the segmentation and detection results for projecting the 2D information to 3D and extracting $z_l^\prime$ by using one-dimensional clustering. The collected 3D information is re-projected to 2D and display the potential hazards.}
\label{fig:software_pipeline}
\end{figure*}

\subsection{Concept}
\label{sec:sys_dgn_concept}
As not too much recent safety management research has involved the safety monitoring of \ac{mic} lifting from bird's-eye view. The application of \ac{ai} in the \ac{lidar}-vision fusion is not explored thoroughly. Thus we provide a few frequently-referred definitions for the easy design of the \ac{top} as well as the potential guidance to the future related work.

\subsubsection{Safety definition and challenges}
\label{sec:sys_dgn_concept_def}
Defining the safety requirements is complex, and it is challenging to consider an all-encompassing guarantee of the construction operators' safety, especially on the construction top during the lifting of the \ac{mic}. \ac{mic} is a free-standing construction component that is delivered from the prefabrication factory to the construction site \cite{tong2024mic}, ensuring an efficient and solid construction manner. The lifting of the \ac{mic} by tower crane is characterized as 
\begin{itemize}
    \item the same starting point of the lifting operation for each building,
    \item the consistent genres of \ac{mic} throughout all construction process, and 
    \item different types of \ac{mic} constructed with the similar structure or material.
\end{itemize}
The autonomous lifting of \ac{mic} is therefore attracting the attention of the construction industry. According to the visualization in Fig.~\ref{fig:overall} (b), we define the space between the tower crane and constructed building as the construction top, where the humans operate and finish the installation of \ac{mic}s. The lifting is achieved by the tower crane with a \ac{mic} frame and the space around the \ac{mic} with a predefined safety tolerant distance is denoted as lifting area. Herein we propose two important safety issues, i.e., protection of the humans under \ac{mic} or aside given the safety tolerant distance and avoidance of the collision between the obstacles, \ac{mic} frames, and cranes given a collision tolerant distance. In spite of other potential safety threatens, we choose these two issues in response to the most concerns from the construction industry.

\subsubsection{Coordinate system}
\label{sec:sys_dgn_concept_coor}
For explicit mathematical representations of the pipeline, as visualized in Fig.~\ref{fig:overall} (a) and (b), four coordinate systems are defined for \ac{top}, including the world coordinate system $\mathbb{W}$, \ac{lidar} coordinate system $\mathbb{L}$, camera coordinate system $\mathbb{C}$, and pixel coordinate system $\mathbb{P}$. The coordinates in the coordinate systems can be transformed mutually via the transformation matrixes, i.e., $^wT_l$ from $\mathbb{L}$ to $\mathbb{W}$, $^lT_c$ from $\mathbb{C}$ to $\mathbb{L}$, and $K$ from $\mathbb{P}$ to $\mathbb{C}$ (known as intrinsic matrix). $\mathbb{W}$ and $\mathbb{L}$ share the same origin and $^wT_l$ only consists of a rotation matrix, i.e., $^wT_l = \begin{bmatrix}
^wR_l & 0 \\
0 & 1
\end{bmatrix}$. A pixel in $\mathbb{P}$, $[x_p, y_p, 1]^T$, can be transferred to a point in $\mathbb{W}$, $[x_w, y_w, z_w]^T$, as follows,
\begin{equation}
\begin{gathered}
\begin{bmatrix}
x_w \\
Y_w \\
z_w \\
1
\end{bmatrix} = \hspace{0.2em} ^wT_l 
\begin{bmatrix}
x_l\\
y_l\\
z_l\\
1
\end{bmatrix} = \hspace{0.2em} ^wT_l \hspace{0.2em} ^lT_c 
\begin{bmatrix}
x_c\\
y_c\\
z_c\\
1
\end{bmatrix}, \\
\begin{bmatrix}
x_c\\
y_c\\
z_c
\end{bmatrix} = K 
\begin{bmatrix}
x_p \cdot z_c\\
y_p \cdot z_c \\
z_c
\end{bmatrix} = 
\begin{bmatrix}
f_x & 0 & c_x\\
0 & f_y & c_y\\
0 & 0 & 1
\end{bmatrix}
\begin{bmatrix}
x_p \cdot z_c\\
y_p \cdot z_c \\
z_c
\end{bmatrix}
\end{gathered}
\label{equ:transformation}
\end{equation}
where $f_x, f_y$ are the scaled focal length in $x$ and $y$ axis respectively, $c_x, c_y$ are the pixel translation in $x$ and $y$ axis respectively, and $z_c$ is unknown by the image in $\mathbb{P}$. Therefore, the \ac{top} pipeline needs to involve the search of $z_c$, as detailed in Section \ref{sec:sys_dgn_soft_framew}. According to the defined safety concerns in Section \ref{sec:sys_dgn_concept_def}, the positions of \ac{mic}s, \ac{mic} frames, and humans at each timestamp in $\mathbb{W}$ are to be localized for a real-time safety monitoring based on our pipeline.

\subsubsection{Point cloud angle distance}
\label{sec:sys_dgn_concept_third}
\begin{figure}
\centering
\includegraphics[width=0.55\textwidth]{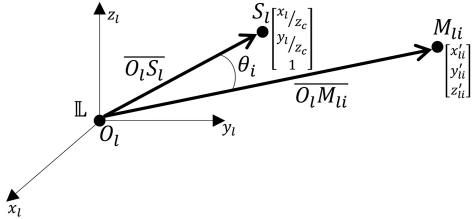}
\caption{The diagram explaining the definition of point cloud angle distance, as mentioned in Section \ref{sec:sys_dgn_concept_third}. $\overline{O_lS_l}$ is the projected vector in $\mathbb{L}$ from a 2D image point in $\mathbb{P}$, and $M_{li}$ is a 3D point in $\mathbb{L}$.}
\label{fig:interior_angle_crop}
\end{figure}

A 2D point in the image space ($\mathbb{P}$) corresponds to a vector in the \ac{pc} data space ($\mathbb{L}$). In the process of \ac{lidar}-vision data fusion, it is necessary to find the closest 3D point in $\mathbb{L}$ to this vector. Thus, we introduce an angle distance metric to measure their distance. Let $\overline{O_lS_l}$ represent the projected vector in $\mathbb{L}$ from a 2D image point in $\mathbb{P}$, the target closet point to $\overline{O_lS_l}$ is denoted as $M_l$, and $M_{li}$ is one of the candidate 3D closest points in \ac{pc} data space ($\mathbb{L}$)\footnote{The definition of angle distance metric is used to measure the distance between a vector $\overline{O_lS_l}$ and a point $M_{li}$ in $\mathbb{L}$, which is to be used in the search of $z'_l$ as mentioned in Section \ref{sec:sys_dgn_soft_framew}.}. The details are shown in Fig.~\ref{fig:interior_angle_crop}, where two auxiliary lines are defined, i.e., $\overline{O_lS_l}$ (from the origin $O_l$ to the point $S_l$) and $\overline{O_lM_{li}}$ (from $O_l$ to the point $M_{li}$). The interior angle $\theta_i$ between these two lines is defined as the angle distance between $\overline{O_lS_l}$ and $M_{li}$. Consequently, the point $M_{li}$ with the lowest angle distance, i.e., $M_l$, indicates the closest point to $\overline{O_lS_l}$ in $\mathbb{L}$.


\subsection{Hardware design}
\label{sec:sys_dgn_hard}
\subsubsection{Components of hardware}
\label{sec:sys_dgn_hard_compn}
The hardware system of \ac{top} contains a host computer (ZOTAC MINI PC ERP74070C), a display monitor, a wireless internet module, and a sensor unit with two sensors, i.e., a remote camera (MV-CS200-10GC) and a \ac{lidar} (Livox AVIA with a built-in \ac{imu}), as visualized in Fig. \ref{fig:overall}. The devices are installed to rotate together with the crane cockpit, ensuring a possible 360-degree view of the construction top space. Importantly, the host computer processes the captured data by the sensor unit locally by using intelligent workflow (discussed in Section \ref{sec:sys_dgn_soft}). The real-time monitoring status of the tower crane operation is displayed in the cockpit for the crane driver. Also, the resulting output is transmitted to the off-site office via wireless internet module for backup and guarantee, thereby releasing the data burden between on-site and off-site devices of the \ac{top}.

\subsubsection{LiDAR-camera calibration}
\label{sec:sys_dgn_hard_cali}
The \ac{lidar}-vision data fusion aims to achieve the information transformation between image collected by camera and \ac{pc} collected by \ac{lidar}. The calibration between them is the key to the task, including the intrinsic ($K$) and extrinsic ($^lT_c$)  calibrations. The calibration procedure of intrinsic follows the toolbox \cite{heikkila1997four,scaramuzza2006toolbox}, which is conducted prior to the calibration of extrinsic transformation matrixes. As the \ac{lidar} used in the system integrates a \ac{imu} that provides the rotation of the \ac{lidar} with respect to $\mathbb{W}$. Therefore, only $^lT_c$ needs calibrating in our system. We execute the extrinsic calibration according to \cite{yuan2021pixel} which conducts the \ac{lidar}-camera calibration in the pixel level\footnote{Extrinsic calibration: \url{https://github.com/hku-mars/livox_camera_calib}.}. 

\subsection{Software design}
\label{sec:sys_dgn_soft}
\subsubsection{Software framework}
\label{sec:sys_dgn_soft_framew}
Considering the fusion of the two sensors as well as making the pipeline easy-to-use, we apply a two-dimensional (2D) detection method in the working flow, shown as Fig.~\ref{fig:software_pipeline}. Given the image captured by the camera, the detection method generates 2D bounding boxes, including \ac{mic} and \ac{mic} frame lifted by the tower crane and humans on the construction top. As the detection has the possibility missing the objects, a post-processing step is added according to the previous detection results. Herein we use a Kalman filter \cite{welch1995introduction} to conduct this task since we assume the moving of \ac{mic} is stable by the tower crane. Besides, the 2D bounding boxes of \ac{mic} and \ac{mic} frame may also contain the information of other instances in the scenarios. A segmentation is used to generate a mask of \ac{mic} within the 2D bounding box area. On the other hand, the \ac{pc} data is captured by \ac{lidar} for gathering the third-dimensional information of the objects. We apply the preprocessing (down-sampling and zero-value removal) to the original \ac{pc} data that contains a great amount of redundant points. The 2D bounding boxes of humans and segmentations of \ac{mic} and \ac{mic} frame in $\mathbb{P}$ are projected to $\mathbb{W}$ given the calibrated intrinsic metric ($K$) of the camera and extrinsic transformations ($^lT_c, ^wT_l$) as described in Section \ref{sec:sys_dgn_concept_coor}. In this work, we let the term ``Projection'' indicate that a point is transferred from pixel coordinate system $\mathbb{P}$ to world coordinate system $\mathbb{W}$, and re-projection vice versa. In response to the problem proposed in Section \ref{sec:sys_dgn_concept_coor}, i.e, $z_c$ yet determined, we rewrite $^lT_c$ as 

\begin{equation}
^lT_c =
\begin{bmatrix}
^lR_c & ^lt_c \\
0 & 1
\end{bmatrix} = \begin{bmatrix}
r_{11} & r_{12} & r_{13} & t_x\\
r_{21} & r_{22} & r_{23} & t_y\\
r_{31} & r_{32} & r_{33} & t_z\\
0 & 0 & 0 & 1
\end{bmatrix}
\label{equ:Tcl_rewrite}
\end{equation} 

and Eq.~\ref{equ:transformation} as

\begin{equation}
\frac{x_c}{z_c} = f_x x_p + c_x, \quad \frac{y_c}{z_c} = f_y y_p + c_y
\label{equ:K_rewrite}
\end{equation}

\begin{equation}
\frac{z_l}{z_c} = r_{31}\frac{x_c}{z_c} + r_{32}\frac{y_c}{z_c} + r_{33} + t_z
                = r_{31} (f_x x_p + c_x) + r_{32} (f_y y_p + c_y) + r_{33} + t_z
\label{equ:Tcl_rewrite_separate}
\end{equation}

The point $[x_p, y_p]^T$ in $\mathbb{P}$ is transferred to ``scaled point'' $[\frac{x_c}{z_c}, \frac{y_c}{z_c}, 1]^T$ in $\mathbb{C}$, denoted as $S_c$, and then to $[\frac{x_l}{z_c}, \frac{y_l}{z_c}, 1]^T$ in $\mathbb{L}$, denoted as $S_l$. We define a line from the origin of $\mathbb{L}$ to $S_l$, denoted as $\overline{O_lS_l}$, and the search of $z_c$ is conducted along with $\overline{O_lS_l}$. Although the 2D image cannot provide the third-dimensional information, \ac{lidar} provide 3D positions in $\mathbb{L}$, and owing to the feature of scanning mode used by \ac{lidar}, only one 3D point is present in each direction \cite{raj2020survey}. Thus, an intuitive way searching $z_c$ is to find the closest point, defined as $M_l$, $[x_l^\prime, y_l^\prime, z_l^\prime]^T$ to the line $\overline{O_lS_l}$ in $\mathbb{L}$. Specifically, the closest point $M_l$ indicates the line $\overline{O_lM_l}$ between the origin and $M_l$ has the smallest interior angle distance with $\overline{O_lS_l}$. More details are described in Section \ref{sec:sys_dgn_concept_third}. According to the find of $M_l$ and corresponding $S_l$, $z_c$ is calculated by solving  
\begin{equation}
\frac{z_l^\prime}{z_c} = r_{31} (f_x x_p + c_x) + r_{32} (f_y y_p + c_y) + r_{33} + t_z 
\label{equ:z_c_solve}
\end{equation}
since $r_{31} (f_x x_p + c_x) + r_{32} (f_y y_p + c_y) + r_{33} + t_z $ is known given a specific point $[x_p, y_p]^T$ in $\mathbb{P}$ and transformation matrices ($^lT_c$ and $K$). As only one closest point $M_l$ is selected to compute $z_c$, possibly with inaccuracy caused by \ac{lidar} error, we propose to use all possible points, denoted as $\{M_{li}\}_{i=0, 1, 2, ...}$, in $\mathbb{L}$ by transferring all the pixels within segments in $\mathbb{P}$ to $\mathbb{L}$, and apply a one-dimensional clustering on all the possible $M_{li}$ to obtain a more accurate $z_c$. As details visualized in Fig.~\ref{fig:software_pipeline}, $z_c$ values for humans, \ac{mic}, and \ac{mic} frames are clustered and transformed to the camera coordinate to ascertain $x_c, y_c$. Based on the 3D information above that is to be transferred to $W$, the dangers are diagnosed by
\begin{itemize}
    \item comparison between the detected humans and \ac{mic} for finding the presence of humans in danger, and
    \item comparison between the detected \ac{mic} frames for exploring the dangers of collision between two lifting processes.
\end{itemize}
Based on the fixed hardware design, the performance of the system mainly relies on the detection and segmentation method, especially for the different types of \ac{mic}s and humans in a tiny size. Thus, we consider comparing different \ac{sota} to find the suitable methods and modify them to our pipeline.

\subsubsection{Methods used in the framework}
\label{sec:sys_dgn_soft_methods}
The section details the choice of the methods for the \ac{top} pipeline. The base of a detection method consists of the feature extraction backbone and bounding box generation head. This work explores the application of YOLO-structure \cite{jocher2022ultralytics,sohan2024review} in the detection of large (\ac{mic} and \ac{mic} frame) and tiny objects (human) simultaneously, as such neural network exhibits strong performance in detection while meeting the real-time requirements. The samples of these objects are unbalanced, i.e., numerous humans and a few \ac{mic}s (\ac{mic} frames). Therefore, both YOLOv5 and YOLOv8 are compared for selecting a sufficient model, as the \ac{dfl} used in YOLOv8 aims to solve the imbalance existing in the data. Besides, a segmentation model is added to fine-tune the detection result, and herein we use the \ac{sam} \cite{kirillov2023segment} to conduct the 2D segmentation by inputting the 2D bounding boxes from YOLO-based detection as the prompt of \ac{sam}. In terms of 3D information processing, two clustering methods, K-Means \cite{arthur2006slow} and Mean-Shift \cite{comaniciu2002mean}, are compared to be used in our pipeline. More comparison and selection of the mentioned models are further discussed in Section \ref{sec:imple_res}.

\begin{figure}
\centering
\includegraphics[width=0.65\textwidth]{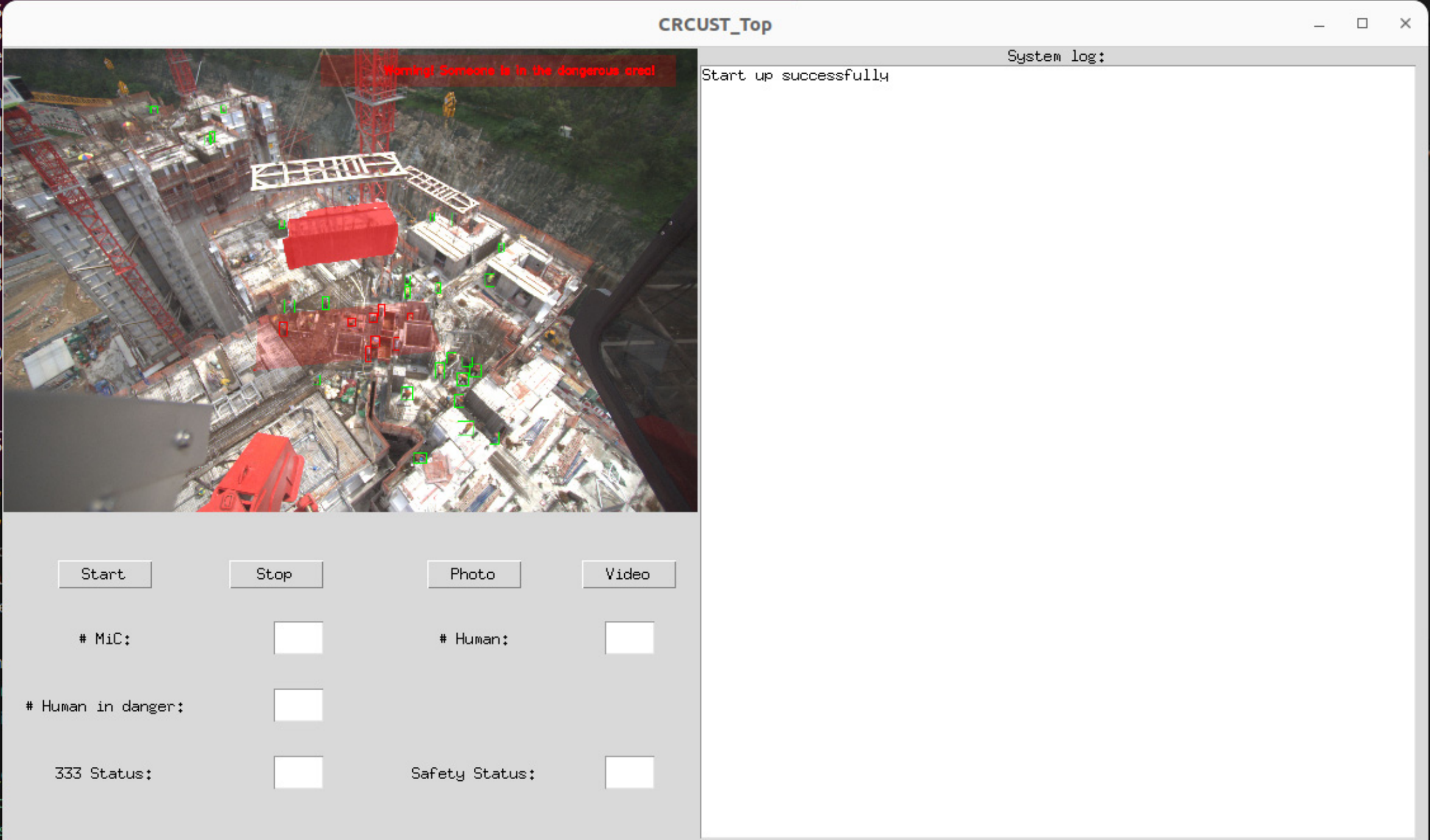}
\caption{The \ac{gui} of \ac{top}, with the real-time display of the safety monitoring, control buttons, and system information logs.}
\label{fig:gui}
\end{figure}

\subsection{Graphical user interface}
\label{sec:sys_dgn_gui}
The proposed \ac{top} pipeline naturally needs a display (\ac{gui}) to warn the crane operators or other personnel on site when dangers occur, as layout in Fig. \ref{fig:gui}. The \ac{gui} presents three components, real-time display of the safety monitoring, essential information regarding safety issues, and control buttons. We display the safe timestamps in green and dangerous timestamps in red with a warning message and alarm. The button contains start and stop of the system plus photo and video capturing capability for back up. Also, the number of \ac{mic} and human workers presented in each frame are listed meanwhile.

\section{Implementation and results}
\label{sec:imple_res}
Given the detailed description of the system design, this section explains the parameters used in the system, method implementation, data collection, model training and comparison, ablation study, as well as results and discussion.

\subsection{Data collection and training}
\label{sec:imple_res_dataset}
For our specific detection task, a customized dataset is one of the core challenges for the pipeline. As 3D information is utilized in the \ac{top} pipeline as visualized in Fig. \ref{fig:software_pipeline}, images and \ac{pc} data were gathered in pairs to prepare our dataset. With the installation of the hardware system (more details in Section \ref{sec:sys_dgn_hard_compn}), we collected 858 image-\ac{pc} pairs for lifting 5 different types of \ac{mic}s and labeled 3 classes in each image, named as \ac{top} dataset, and the statistics of the dataset are outlined in Table \ref{table:crcust}. All data in the dataset were collected on a real site (Public Housing Developments at constructions sites R2-6 and R2-7 on Anderson Road Quarry, Hong Kong SAR.) We split the dataset into train (482 images embodying 11388 human workers, 472 \ac{mic}s, and 593 \ac{mic} frames) and test (376 images presenting 10673 humans workers, 312 \ac{mic}s, and 505 \ac{mic} frames) sets. Different categories of the \ac{mic}s were considered in train and test sets to examine the generalizability of the detection models. Also, the resolution of the images and sampling ability of the \ac{lidar} are summarized in Table \ref{table:crcust}, with further details describing the sizes of the bounding boxes of humans, \ac{mic}s, and \ac{mic} frames shown in the images. The statistics mentioned above present three challenging facts in the pipeline in terms of the related objects, i.e., imbalance of the number, diversities of the sizes, and changing scales of the same class. To this end, the employed method is required to migrate the problems for fulfilling a sufficient pipeline on the construction site. The training of the detection models was executed on the computer (AMD\circledR  Ryzen 9 5959x) with a RTX4090 D X3 24GB until the convergence of the models, and as mentioned in Section \ref{sec:sys_dgn_hard_compn}, the host computer used in on the real site has an i7-14700 processor with a RTX4070 Super 12 GB. The average time of computation used for each frame is 1.639 s, including detection, segmentation, projection, and visualization. 

\begin{table}[h]
\caption{The statistics of the \ac{top} dataset.}
\label{table:crcust}
\centering
\small
\renewcommand{\arraystretch}{1.25}
\begin{tabular}{@{}llll@{}}
\toprule
\multicolumn{2}{@{}l@{}}{Split} &
Train & Test \\
\midrule 
\multicolumn{2}{@{}l@{}}{\makecell[l]{Number of pairs:\\images and point clouds}} & 482 & 376\\
\midrule
\multicolumn{2}{@{}l@{}}{Number of humans} & 11388 & 10673 \\
\midrule
\multicolumn{2}{@{}l@{}}{Number of \ac{mic}s} & 472 & 312 \\
\midrule
\multicolumn{2}{@{}l@{}}{Number of \ac{mic} frames} & 593 & 505 \\
\midrule
\multicolumn{2}{@{}l@{}}{\makecell[l]{Image resolution\\(width $\times$ height, pixels)}}  & \multicolumn{2}{l}{$5472\times3648$} \\
\midrule
\multicolumn{2}{@{}l@{}}{\makecell[l]{Number of points for\\each \ac{pc} frame}} &  \multicolumn{2}{l}{24000} \\
\midrule
& Human & $73.83 \times 91.29$ & $73.76 \times 88.47$ \\
\cmidrule{2-4}
& \ac{mic} & $467.16 \times 702.14$ & $394.18 \times 427.47$ \\
\cmidrule{2-4}
\makecell[l]{Average size of bounding boxes:\\width $\times$ height (pixels)}   & \ac{mic} frame & $592.11 \times 639.40$ & $789.87 \times 446.50$ \\
\bottomrule
\end{tabular}
\normalsize
\end{table}


\begin{figure}[h!]
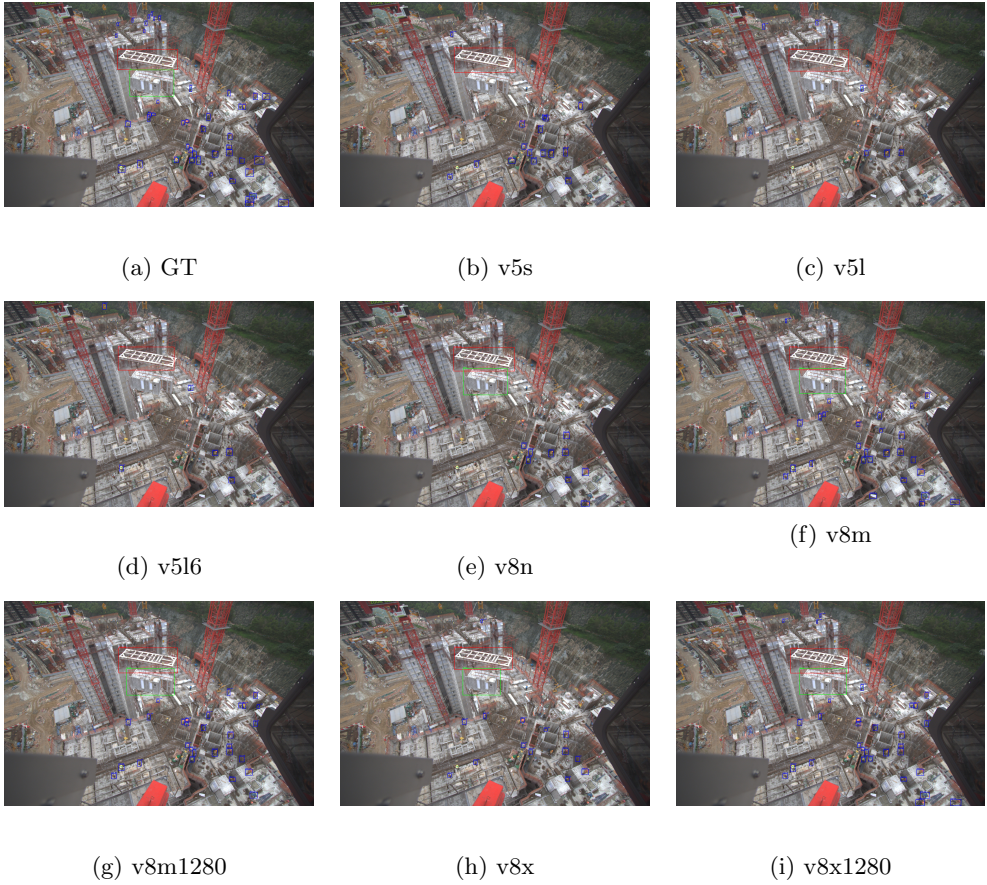

\centering
    \begin{subfigure}[t]{1.6in}
    \includegraphics[width=1.6in]{res_gt_camera_150.png.VwzhMSASfI87g2HnfrSU-eps-converted-to.pdf}
    \label{fig:gt_camera_150}
    \caption{GT}
    \end{subfigure}
    ~
    \centering
    \begin{subfigure}[t]{1.6in}
    \centering
    \includegraphics[width=1.6in]{res_v5s_camera_150.png.VwzhMSASfI87g2HnfrSU-eps-converted-to.pdf}
    \label{fig:v5_s_camera_150}
    \caption{v5s}
    \end{subfigure}
    ~
    \centering
    \begin{subfigure}[t]{1.6in}
    \centering
    \includegraphics[width=1.6in]{res_v5l_camera_150.png.VwzhMSASfI87g2HnfrSU-eps-converted-to.pdf}
    \label{fig:v5_l_camera_150}
    \caption{v5l}
    \end{subfigure}
    ~
    \centering
    \begin{subfigure}[t]{1.6in}
    \centering
    \includegraphics[width=1.6in]{res_v5l6_camera_150.png.VwzhMSASfI87g2HnfrSU-eps-converted-to.pdf}
    \label{fig:v5_l6_camera_150}
    \caption{v5l6}
    \end{subfigure}
    ~
    \centering
    \begin{subfigure}[t]{1.6in}
    \centering
    \includegraphics[width=1.6in]{res_v8n_camera_150.png.VwzhMSASfI87g2HnfrSU-eps-converted-to.pdf}
    \label{fig:v8_n_camera_150}
    \caption{v8n}
    \end{subfigure}
    ~
    \centering
    \begin{subfigure}[t]{1.6in}
    \centering
    \includegraphics[width=1.6in]{res_v8m_camera_150.png.VwzhMSASfI87g2HnfrSU-eps-converted-to.pdf}
    \caption{v8m}
    \end{subfigure}
    ~
    \centering
    \begin{subfigure}[t]{1.6in}
    \centering
    \includegraphics[width=1.6in]{res_v8m1280_camera_150.png.VwzhMSASfI87g2HnfrSU-eps-converted-to.pdf}
    \label{fig:v8_m_1280_camera_150}
    \caption{v8m1280}
    \end{subfigure}
    ~
    \centering
    \begin{subfigure}[t]{1.6in}
    \centering
    \includegraphics[width=1.6in]{res_v8x_camera_150.png.VwzhMSASfI87g2HnfrSU-eps-converted-to.pdf}
    \label{fig:v8_x_camera_150}
    \caption{v8x}
    \end{subfigure}
    ~
    \centering
    \begin{subfigure}[t]{1.6in}
    \centering
    \includegraphics[width=1.6in]{res_v8x1280_camera_150.png.VwzhMSASfI87g2HnfrSU-eps-converted-to.pdf}
    \label{fig:v8_x_1280_camera_150}
    \caption{v8x1280}
    \end{subfigure}
\caption{The comparison of the detection results by YOLOv5 and YOLOv8 with IoU threshold = 0.3, Confidence threshold = 0.5 (humans), and Confidence threshold = 0.7 (\ac{mic}s and \ac{mic} frames).}
\label{fig:vis_detect_res_2}
\end{figure}

\subsection{Evaluation metric}
\label{sec:imple_res_eval}
Two groups of methods need comparisons for the selection of the models, i.e., detection model and one-dimensional clustering, so we considered \ac{map} \cite{padilla2020survey} and distance error ($err$). The definition of \ac{map} is based on the \ac{iou} that measures the positional similarity of two bounding boxes. Given two bounding boxes, $B_p$ (prediction) and $B_g$ (ground-truth), the \ac{iou} between them is computed as 

\begin{equation}
IoU = \frac{A(B_p \cup B_g)}{A(B_p \cap B_g)}
\label{equ:precision}
\end{equation}
where $A(B_p \cup B_g)$ is the area of the overlap between $B_p$ and $B_g$, and $A(B_p \cap B_g)$ is the area of the union between $B_p$ and $B_g$.
 
The precision ($P$) and recall ratio of an object detection model can be computed as

\begin{equation}
P = \left.\frac{TP}{TP + FP} \right|_{T_{IoU}, T_c}
\label{equ:precision}
\end{equation}

\begin{equation}
R = \left.\frac{TP}{TP + FN} \right|_{T_{IoU}, T_c}
\label{equ:precision}
\end{equation}
where TP, FP, and FN indicate
\begin{itemize}
    \item True Positive (TP): the correct detection with confidence value over $T_c$, i.e., its IoU with corresponding ground truth over a threshold $T_{IoU}$;
    \item False Positive (FP): the incorrect detection with confidence value over $T_c$, i.e., its IoU with wrongly matched ground truth over a threshold $T_{IoU}$;
    \item False Negative (FN): the undetected ground-truth with confidence value over $T_c$, i.e., its IoU with all detection lower than $T_{IoU}$.
\end{itemize}
In practice, $P$ and $R$ can be displayed as a trade-off curve given different $T_c$. Average Precision ($AP$) is usually used to combine these two indexes by integrating the curve from $T_c = 0$ to $T_c = 1.0$. Moreover, \ac{map} is the mean of $AP$ over all classes in dataset. We use two values (50\% and 95\%) of $T_{IoU}$ to test the model, i.e., mAP50 and mAP95.

Additionally, we aim to measure the localization accuracy in 3D, so the error distance between two 3D points is computed as 

\begin{equation}
err = \sqrt{(x' - x_g)^2 + (y' - y_g)^2 + (z' - z_g)^2}
\label{equ:precision}
\end{equation}
where $\{x', y', z'\}$ is the predicted point, and $\{x_g, y_g, z_g\}$ is the ground-truth point.

\subsection{Comparison of detection models}
\label{sec:imple_res_comp}
Although the \ac{sota} of the detection methods have emerged as a success, many challenges in our safety case cannot still be solved. Thus, 6 YOLOs were compared on our customized data by using the mAP50 and mAP50-95 (the higher the better), as visualized in Table \ref{table:comparison} and in Fig. \ref{fig:vis_detect_res_2}. Among them, two image sizes (640 and 1280) were considered in YOLOv8m and YOLOv8x. The results delivered that YOLOv8x outperformed the other models in most indexes, especially in image size at 1280. Nonetheless, the size of the YOLOv8x was almost three times larger than YOLOv8m, which, at the meantime, had comparable accuracy. Therefore, YOLOv8m was taken into account in the pipeline.

\subsection{Safety monitoring visualization}
\label{sec:imple_res_sf_vis}
\begin{figure}[h!]
\centering
    \begin{subfigure}[t]{2.2in}
    \centering
    \includegraphics[width=2.2in]{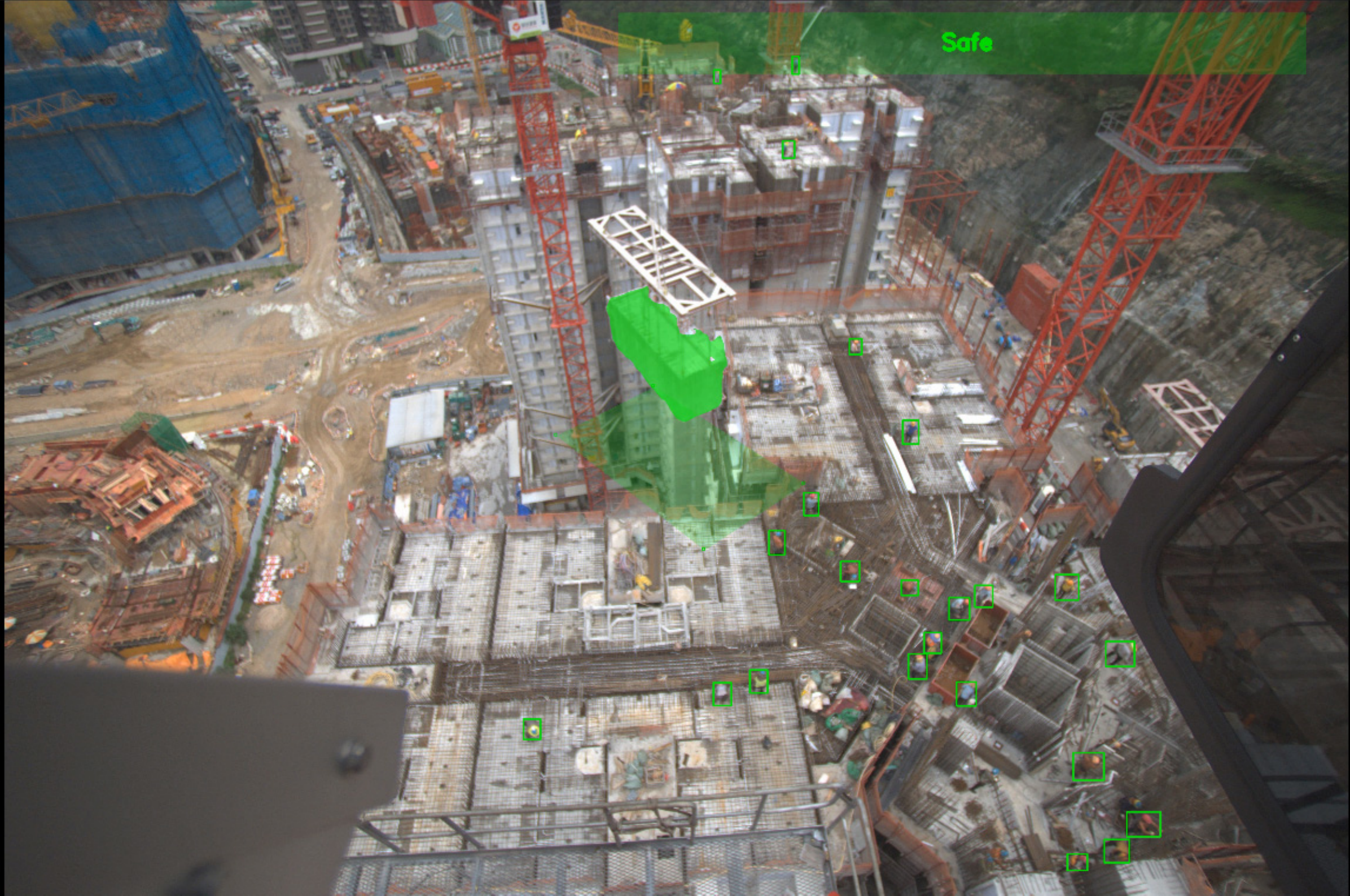}
    \caption{}
    \label{fig:vis_safe}
    \end{subfigure}
    ~
    \begin{subfigure}[t]{2.2in}
    \centering
    \includegraphics[width=2.2in]{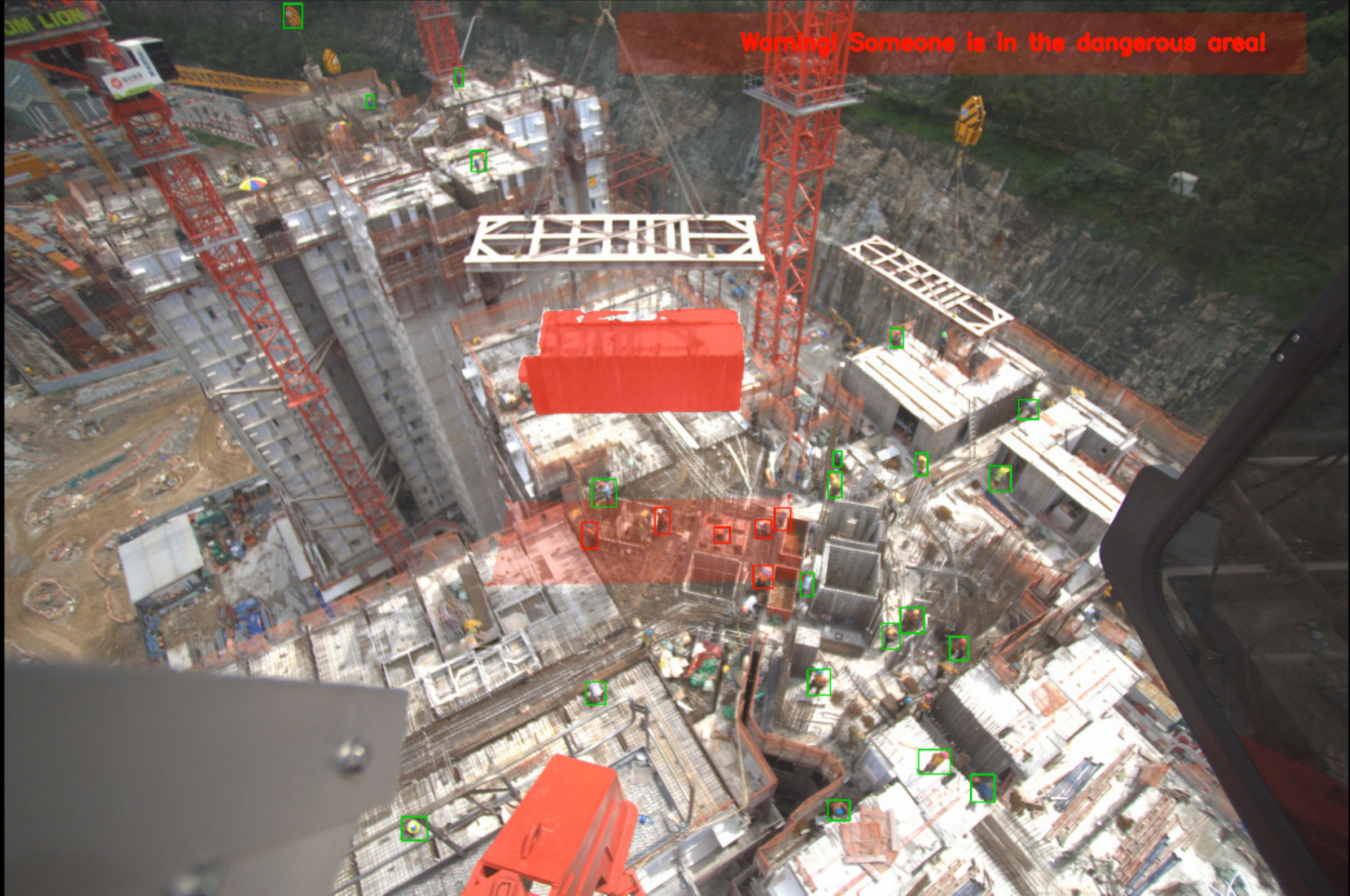}
    \caption{}
    \label{fig:vis_danger}
    \end{subfigure}
    ~
    \begin{subfigure}[t]{2.2in}
    \centering
    \includegraphics[width=2.2in]{vis_safe2-eps-converted-to.pdf}
    \caption{}
    \label{fig:vis_safe2}
    \end{subfigure}
    ~
    \begin{subfigure}[t]{2.2in}
    \centering
    \includegraphics[width=2.2in]{vis_danger2-eps-converted-to.pdf}
    \caption{}
    \label{fig:vis_danger2}
    \end{subfigure}
\caption{The visualization of displaying the safe (green) and dangerous (red) cases during the lifting of two types of \ac{mic}s.}
\label{fig:vis_safe_danger}
\end{figure}

The safe monitoring intuitively requires a display for warning the tower crane operator in case of dangers. We simplified the display by using two colors showing safe and dangerous cases of lifting the \ac{mic}, i.e., without and with danger during the moving of \ac{mic}. As visualized in Fig.~\ref{fig:vis_safe_danger}, the safe case is colored in green (Fig.~\ref{fig:vis_safe} and ~\ref{fig:vis_safe2}) and the dangerous case in red (Fig.~\ref{fig:vis_danger} and ~\ref{fig:vis_danger2}). The 3D area of the \ac{mic} was enlarged with a safety tolerance and re-projected back to the mask in the image space. Any human presenting in the mask or collision triggered the warning of dangers. The projection plane was the same as the construction top, where the human operators were working.

\subsection{Ablation study}
\label{sec:imple_res_comp_ablation}
The \ac{top} pipeline integrated various methods to take the mutual
advantages, so the lack of any specific one can undermine the pipeline's performance. To verify the functions of the methods involved in the pipeline, we conducted the ablation study for the segmentation part and one-dimensional clustering.

\begin{figure}[h!]
\centering
\includegraphics[width=0.75\textwidth]{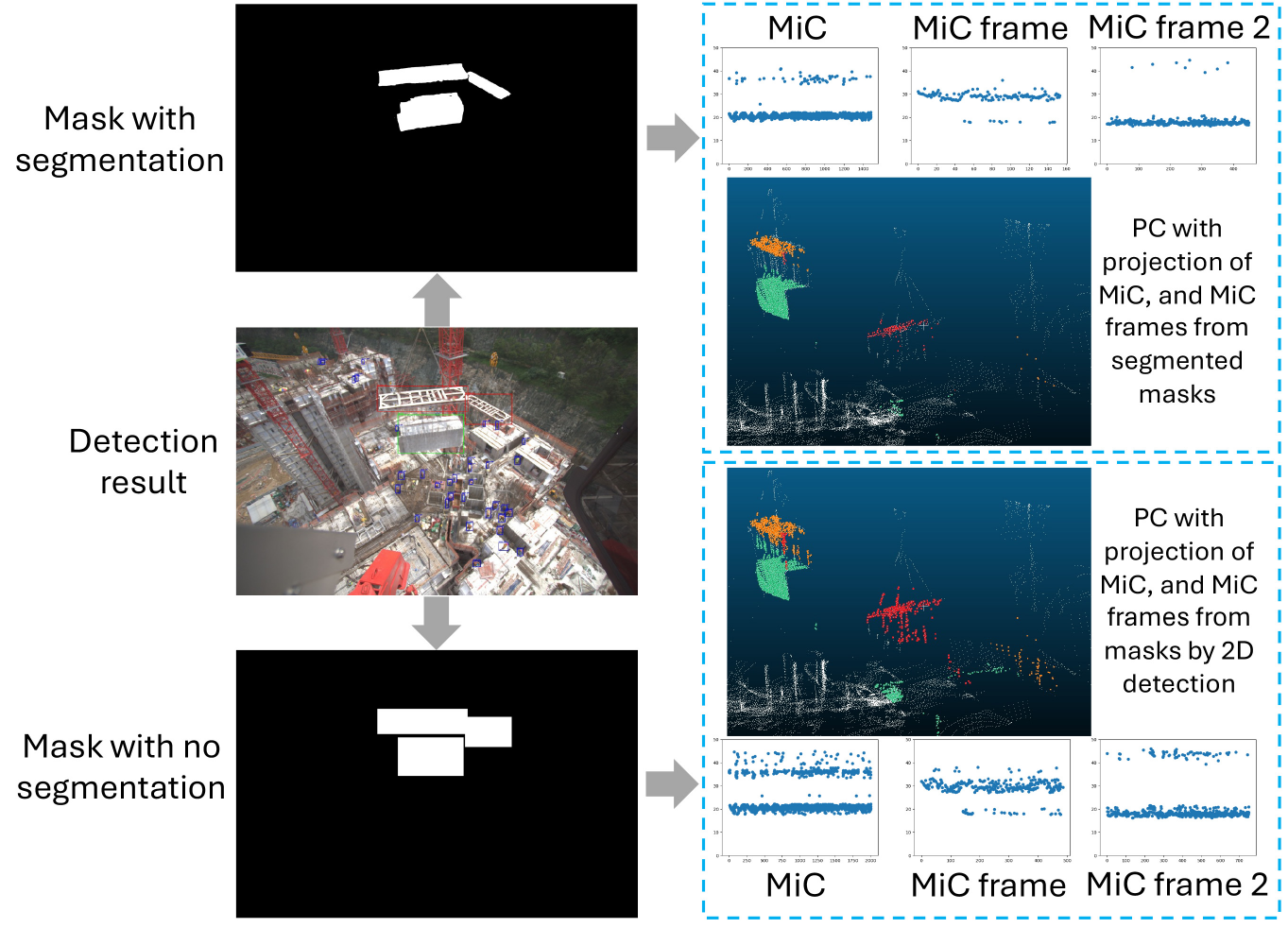}
\caption{The comparison for $z_l^\prime$ extraction with and without segmentation. In \ac{pc} data, \ac{mic} points are in green, one \ac{mic} frame in yellow, and another in red.}
\label{fig:comparison_seg}
\end{figure}

\textbf{Segmentation} A \ac{sam} was applied to the detected 2D bounding boxes for generating fine object instance areas of \ac{mic} and \ac{mic} frame. We compared two projections of the object areas in the 3D \ac{pc} data, i.e., from masks generated by \ac{sam} and from these generated merely by the detection result (2D bounding boxes). As detailed in Fig.~\ref{fig:comparison_seg}, the projection with only detection result caused more noise points in \ac{pc} data, thereby consequently extracted $z_l^\prime$ presenting less obvious clusters, compared with that by \ac{sam}. The core function of the \ac{sam} was to remove the background area within the mask and then reduced the noise points in terms of extracting $z_l^\prime$. An accurate $z_l^\prime$ searching can ensure a better localization of the target objects (\ac{mic} and \ac{mic} frame).

\begin{table}[!h]
\caption{The comparison of clustering methods for obtaining $z_l^\prime$.}
\label{table:comp_clustering}
\centering
\small
\renewcommand{\arraystretch}{1.25}
\begin{tabular}{c c c c}
\toprule
Method & Averaging & K-Means & Mean-Shift \\
\midrule
$err$ (m) & 4.12 & 1.41 & 3.58 \\
\bottomrule
\end{tabular}
\normalsize
\end{table}

\textbf{One-dimensional clustering} As the third dimensional information of the target objects was essential to the safety monitoring, a proper selection among the abundant extracted $z_l^\prime$ by \ac{sam} had a decisive role in the 3D localization. We aimed to make the pipeline less computational and easy-to-use, so three simple but sufficient methods were compared in Table~\ref{table:comp_clustering}. In order to evaluate the accuracy of the clustering in our case, 8 objects (\ac{mic} and \ac{mic} frame) in different \ac{pc} data from the dataset were randomly selected with the corresponding centers chosen manually as ground truth. We compared the calculation of the 3D localization by three methods, i.e., averaging, K-Means, and Mean-Shift for $z_l^\prime$, with the ground truth and Table \ref{table:comp_clustering} outlines the distance errors ($err$). The K-Means clustering outperformed the others and had less computation than Mean-Shift as well, so it was implemented in our \ac{top} pipeline. 

\begin{sidewaystable}
\caption{The comparison of the YOLOv5 and YOLOv8 regarding model size, mAP50, and mAP50-95.}
\label{table:comparison}
\begin{tabular*}{\textheight}{@{\extracolsep\fill}lccccccccccc}
\toprule
&  & &  & \multicolumn{4}{@{}c@{}}{mAP50} & \multicolumn{4}{@{}c@{}}{mAP50-95} \\ \\
\cmidrule{5-8}\cmidrule{9-12}
Model & \makecell{Batch\\size} & \makecell{Image\\size} & \makecell{Model size\\(params)} & Human &  \ac{mic} & \ac{mic} frame & All & Human &  \ac{mic} & \ac{mic} frame & All \\
\midrule
YOLOv5s &16 &640 &7.2M &0.418 &0.218 &0.941 &0.526 &0.128 &0.107 &0.584 &0.273 \\ 
\midrule
YOLOv5l &16 &640 &46.5M &0.541 &0.188 &0.91 &0.546 &0.202 &0.113 &0.657 &0.324 \\ 
\midrule
YOLOv5l6 &16 &640 &76.8M &0.428 &0.128 &0.782 &0.446 &0.154 &0.0465 &0.446 &0.215  \\ 
\midrule
YOLOv8n &16 &640 &3.2M &0.493 &0.669 &0.951 &0.705 &0.169 &0.425 &0.72 &0.438 \\ 
\midrule
\multirow{2}{*}{YOLOv8m} &16 &640 &\multirow{2}{*}{25.9M} &0.629 &0.83 &0.985 &0.815 & 0.237 &0.492 &0.784 &0.504 \\
 &2 &1280 & &\textbf{0.807} &0.819 &0.986 &0.871 &\textbf{0.363} &0.523 &0.792 &0.559  \\
\midrule
\multirow{2}{*}{YOLOv8x} &16 &640 &\multirow{2}{*}{68.2M} &0.586 &0.851 &0.985 &0.807 &0.213 &0.514 &\textbf{0.801} &0.509 \\
 &2 &1280 & &0.774 &\textbf{0.887} &\textbf{0.992} &\textbf{0.884} &0.336 &\textbf{0.599} &0.762 &\textbf{0.566}  \\
\bottomrule
\end{tabular*}
\normalsize
\end{sidewaystable}

\subsection{Limitations}
\label{sec:imple_res_comp_limit}
This work presents a system for the safety monitoring of construction top, and in most cases, the dangers can be detected and warned, although some limitations need to be noticed. We heuristically selected a higher confidence threshold for \ac{mic} and \ac{mic} frame detection as the detection method may pick some false positives owing to the limited features and similar feature to the surroundings. However, a higher threshold may result in more missing of the detection. Besides, a lower confidence threshold was applied to the human detection for ensuring no missing of possible humans. The center of the bounding box was used as the position of humans, so a small part of the 'fake' detection of human body occurring in the dangerous area may cause the warning, i.e., false positives. Also, even though the object occlusion did not usually occur in our site case, the occlusion with near distance between two objects shall be avoided when users evaluate our system for use. Last but importantly, the inference time of the whole pipeline needs improving for a better real-time display on site, even though it is sufficient for current safety monitoring.

\section{Conclusion}
\label{sec:conclusion}
In response to the increasing safety concerns on the construction sites, typically those with \ac{mic} involved, we introduce four questions regarding the safety monitoring of \ac{mic} lifting from bird's-eye view. This work tackles the questions by 1) defining two specific safety concerns for tower crane and workforce on the construction top; 2) designing a \ac{top} system that realizes the monitoring of the defined safety concerns with automated warning triggers; 3) fulfilling accurate localization of \ac{mic}, \ac{mic} frame, and human workers by fusing the \ac{sota} 2D object detection with 3D \ac{pc} data collected by \ac{lidar}; 4) installing the proposed system on a real site and collecting a site dataset for verification of the implemented models; 5) additionally, assessing the functions of essential component in the software framework by an ablation study. In summary, \ac{top} holds functions of AI-based detection capability, 3D measurements, danger alert, and display, and can be applied to the real site for the safety monitoring task. The results and visualization demonstrated the ability of the pipeline conducting danger detection and warning (display). Subsequently, we aim to research the potential application of open-vocabulary object detection, so that no real-site datasets are necessary, which is usually a highly challenging step in terms of \ac{ai} application on site. Also, the proposed \ac{top} can be relocated to other sites without further modifications, and a commercialization step is under consideration as well, aiming to improve the safety guarantee on more real-world sites.

\bmhead{Acknowledgements}
The work was funded by InnoHK-HKCRC. The authors would like to thank the project support from HKCRC Limited and data acquisition support on the construction site from Development and Construction Division, Housing Department, Government of Hong Kong, Hong Kong, China. Additionally, the authors also thank the colleagues, Mr. Zhengyao Liu and Mr. Wanyou Yang, for assisting with the hardware design and figure preparation.

\bmhead{Data availability}
Data will be made available on request.

\bmhead{Notes on contributors}
Conceptualization and data collection, Y. Wang, Y. H. Ng, H. Liang, C. W. Chang, H. Chen; methodology, experiments, validation, investigation, and visualization, Y. Wang, Y. H. Ng, C. W. Chang, H. Chen; formal analysis and writing|original draft preparation, Y. Wang, H. Chen; writing|review and editing, Y. Wang, Y. H. Ng, C. W. Chang, H. Chen. All authors have read and agreed to the published version of the manuscript.

\bmhead{Declarations}
The authors declare no conflict of interest.

\bibliography{reference}

\end{document}